\documentclass[sigconf]{acmart}

\usepackage{amsmath}
\usepackage{mathtools} 
\usepackage{subcaption}
\usepackage{booktabs}
\usepackage{adjustbox}
\usepackage{multirow}
\usepackage{multicol}
\usepackage{xcolor}
\usepackage{csquotes}

\usepackage{fancyhdr}

\definecolor{green}{rgb}{0.12, 0.85, 0.6} 

\definecolor{red}{rgb}{0.89, 0.09, 0.29} 

\newcommand{\redpart}[1]{{\color{red}{#1}}}

\usepackage{colortbl}
\definecolor{defaultcolor}{gray}{.9}
\newcommand{\df}[1]{\cellcolor{defaultcolor}{#1}}

\AtBeginDocument{%
  }

\setcopyright{acmcopyright}
\acmYear{2023}
\acmConference[MM '23] {Proceedings of the 31st ACM International Conference on Multimedia}{October 29--November 3, 2023}{Ottawa, ON, Canada.}
\acmBooktitle{Proceedings of the 31st ACM International Conference on Multimedia (MM '23), October 29--November 3, 2023, Ottawa, ON, Canada}
\acmPrice{15.00}
\acmISBN{979-8-4007-0108-5/23/10}
\acmDOI{10.1145/3581783.3612035}

\renewcommand\footnotetextcopyrightpermission[1]{}

\begin{document}

\title{View while Moving: Efficient Video Recognition in Long-untrimmed Videos}

\author{Ye Tian}
\authornote{Both authors contributed equally to this research and share the co-first authorship.}
\affiliation{%
  \institution{State Key Laboratory of Networking and Switching Technology, Beijing University of Posts and Telecommunications}
  \city{Beijing}
  \country{China}
}

\author{Mengyu Yang}
\authornotemark[1]
\authornote{Corresponding authors: mengyuyang@bupt.edu.cn, zls326@bupt.edu.cn.}
\affiliation{%
  \institution{State Key Laboratory of Networking and Switching Technology, Beijing University of Posts and Telecommunications}
  \city{Beijing}
  \country{China}
}

\author{Lanshan Zhang}
\authornotemark[2]
\affiliation{%
  \institution{Beijing Key Laboratory of Network System and Network Culture, Beijing University of Posts and Telecommunications}
  \city{Beijing}
  \country{China}
}

\author{Zhizhen Zhang}
\affiliation{%
  \institution{DCST, Tsinghua University}
  \city{Beijing}
  \country{China}
}

\author{Yang Liu}
\affiliation{%
  \institution{State Key Laboratory of Networking and Switching Technology, Beijing University of Posts and Telecommunications}
  \city{Beijing}
  \country{China}
}

\author{Xiaohui Xie}
\affiliation{%
  \institution{DCST, Tsinghua University}
  \city{Beijing}
  \country{China}
}

\author{Xirong Que}
\affiliation{%
  \institution{State Key Laboratory of Networking and Switching Technology, Beijing University of Posts and Telecommunications}
  \city{Beijing}
  \country{China}
}

\author{Wendong Wang}
\affiliation{%
  \institution{State Key Laboratory of Networking and Switching Technology, Beijing University of Posts and Telecommunications}
  \city{Beijing}
  \country{China}
}

\renewcommand{\shortauthors}{Ye Tian et al.}

\begin{teaserfigure}
\includegraphics[width=\textwidth]{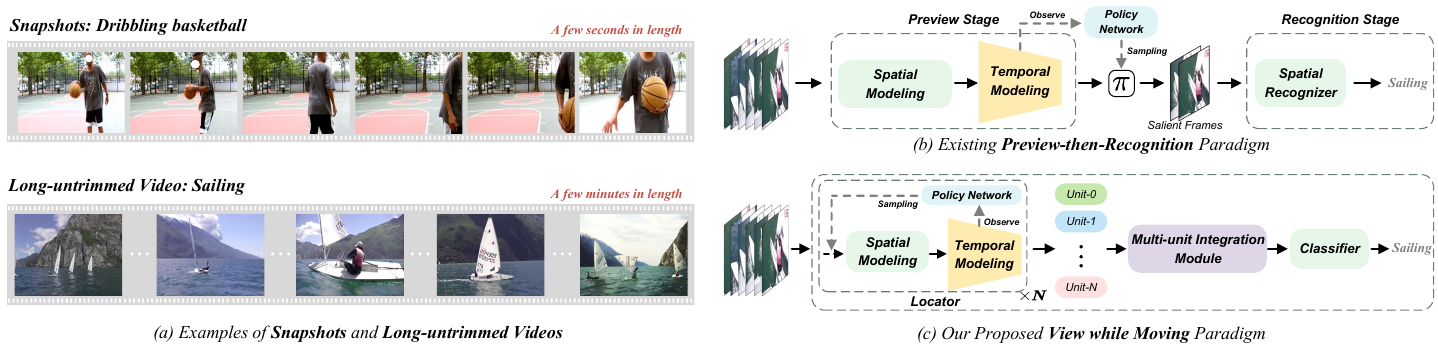}
\caption{(a) Examples of snapshots and long-untrimmed videos: unlike short-trimmed snapshots spanning a few seconds that encompass several actions, long-untrimmed videos generally depict complex activities or stories with numerous transitions of shots and scenes, consisting of multiple semantic units. (b) and (c) Comparison of the existing paradigm and our proposed ViMo: ViMo only accesses the raw frame at once during inference, and has a hierarchical mechanism to capture the unit- and video-level temporal semantics respectively.}
\label{fig1}
\end{teaserfigure}

\begin{abstract}
Recent adaptive methods for efficient video recognition mostly follow the two-stage paradigm of \enquote{preview-then-recognition} and have achieved great success on multiple video benchmarks. 
However, this two-stage paradigm involves two visits of raw frames from coarse-grained to fine-grained during inference (cannot be parallelized), and the captured spatiotemporal features cannot be reused in the second stage (due to varying granularity), being not friendly to efficiency and computation optimization.
To this end, inspired by human cognition, we propose a novel recognition paradigm of \enquote{View while Moving} for efficient long-untrimmed video recognition.
In contrast to the two-stage paradigm, our paradigm only needs to access the raw frame once. 
The two phases of coarse-grained sampling and fine-grained recognition are combined into unified spatiotemporal modeling, showing great performance.
Moreover, we investigate the properties of semantic units in video and propose a hierarchical mechanism to efficiently capture and reason about the unit-level and video-level temporal semantics in long-untrimmed videos respectively.
Extensive experiments on both long-untrimmed and short-trimmed videos demonstrate that our approach outperforms state-of-the-art methods in terms of accuracy as well as efficiency, yielding new efficiency and accuracy trade-offs for video spatiotemporal modeling.
\end{abstract}



\maketitle

\section{Introduction}
Video recognition is a fundamental task in video understanding, having a wide range of application scenarios, such as video recommendation \cite{hmmved,videogamerecommendation}, retrieval \cite{clip4clip,disentangled,actionbytes} and transmission \cite{deepgame}, \textit{etc}. 
Over the past few years, remarkable success \cite{twostream,tsn,lrcn,c3d} has been achieved in this task within the context of deep learning. 
However, to handle multiple frames in videos for spatiotemporal modeling, video models are relatively expensive in terms of complexity and computational cost, constraining their deployment in many resource-limited applications, such as edge computing \cite{edgesegmentation,ekya}, online vision \cite{tevit,du2020server,autocaptiondataset}, \textit{etc}. 
Consequently, it becomes crucial to devise efficient models that preserve recognition accuracy while minimizing the computational expense, especially for long-untrimmed videos with more frames \cite{multiagent,adaframe,listentolook,memvit}.

Given this goal, a number of recent works \cite{fastforward,frameglimpse,multiagent,adaframe,scsampler,liteeval,listentolook,arnet,adamml,videoiq,nsnet,tsqnet} propose to adaptively reduce the temporal redundancy and noise in video recognition. 
As shown in Figure \redpart{\ref{fig1}(b)}, task-relevant salient frames are picked as the identification root, while redundancies and noises are discarded, leading to significant efficiency gains. 
Existing methods mostly rely on a two-stage paradigm of \enquote{preview-then-recognition}, i.e., the adaptive selection is performed before the frames are sent to recognizer \cite{afnet}. 
A coarse-grained spatiotemporal extractor first previews the video and adaptively selects salient frames with a policy network. 
Then, the selected frames are fed into a TSN-like \cite{tsn} fine-grained recognizer, outputting final results relying on spatial features. 
However, this paradigm suffers from two drawbacks: 
(i) The paradigm involves two times of raw frame feature extraction from coarse-grained to fine-grained, which are linear and cannot be parallelized during inference. 
Obviously, it is not friendly to efficiency and computational optimization. 
(ii) The spatiotemporal features captured in the first phase are dropped, and cannot be reused in the second phase (due to varying granularity), resulting in information wastage. 
Also, the second phase lacks the modeling of the temporal semantics, relying only on spatial. 

Moreover, unlike snapshots or clips with only a few actions, the long-untrimmed video consists of multiple semantic units \cite{actionbytes, cmhm}, as shown in Figure \redpart{\ref{fig1}(a)}. A number of works in the past have explored the temporal semantics modeling, using mechanisms such as LSTM \cite{fastforward,adaframe,adamml,adafocusv2}, Pool \cite{scsampler,frameexit}, and self-attention \cite{nonlocal,tsqnet,nsnet} to model the temporal features between frames. 
Although regarding video as a whole or multiple splits is effective for action modeling, these methods ignore the property of semantic units in videos, which is unfavorable for long-untrimmed video.
It is obviously intractable to model the long-term memory when multiple semantic units are mixed together.
Also, as a whole, when video length grows, the observation and decision space of the model becomes unacceptable. 
Temporal modeling turns out to be a computation and memory-intensive task. It is not conducive to scaling to long-untrimmed videos.

To overcome the above drawbacks, we rethink how to achieve efficient video recognition in light of human long-term cognitive habits \cite{humanshortlongtermmemory,ran2021deep}. If asked to understand a long-untrimmed video on the fly, such as a movie, we would quickly browse the key contents of the video and skip the dull parts according to our experience. It is worth noting that we view the contents while moving the progress bar, rather than picking out key parts and then rewatching them again, which saves us a considerable amount of time. In addition, for such long-untrimmed videos containing multiple semantic units, we generally follow a habit of building semantics from the local to the global. Concretely, we first analyze each unit to capture the unit-level semantics. After understanding multiple units, we connect them to reason about the whole story of the video \cite{humanreasoning}. This habit of hierarchical analysis makes it easy to reason about the complex semantics across various levels in the video.

Inspired by human cognition, we introduce a new recognition paradigm of \enquote{\textbf{Vi}ew while \textbf{Mo}ving} (ViMo) for efficient video recognition in long-untrimmed videos. In contrast to the twice visits of the existing two-stage adaptive recognition paradigm, our paradigm ViMo only accesses the raw frame one time. We show that the unified paradigm of \enquote{View while Moving} is more efficient than the two-stage paradigm of \enquote{preview-then-recognition}. 
Moreover, for temporal modeling, we investigate the properties of semantic units in video and propose a hierarchical mechanism to efficiently capture and reason about the unit-level and video-level temporal semantics in long-untrimmed videos, respectively.
Extensive experimental results on ActivityNet \cite{activitynet}, FCVID \cite{fcvid} and Kinetics-Sounds \cite{kinetics-sounds}, demonstrate that our proposed ViMo outperforms state-of-the-art methods in terms of accuracy as well as efficiency.
With only ImageNet-1K pretraining, our ViMo achieves 82.4\% and 86.4\% mAP with 38.7 and 36.4 GFLOPs on ActivityNet and FCVID respectively.
In addition, ViMo also achieves state-of-the-art accuracy of 92.4\%, while saving 76.2\% in GFLOPs on Kinetics-Sounds with short-trimmed videos.

In summary, our main contributions are as follows:
\begin{itemize}
    \item We propose a new \enquote{View while Moving} paradigm for efficient long-untrimmed video recognition. Our new paradigm only accesses the raw frame at once during inference.
    \item We propose a hierarchical mechanism, which adaptively observes the local semantic units and reasons about the video-level temporal semantics for efficient spatiotemporal modeling in long-untrimmed videos.
    \item Extensive experiments conducted on several commonly used benchmarks demonstrate that our ViMo yields improved accuracy-efficiency trade-offs over existing state-of-the-art methods.
\end{itemize}

\section{Related Work}
\subsection{Video Recognition}
Due to its wide range of applications, video recognition has been one of the most active research areas.
Early methods mainly used hand-crafted local features and single-layer classifiers (\textit{e.g.}, LR, SVM, \textit{etc}.) to study video recognition \cite{idt}. 
In the context of deep learning, performed by large public video datasets \cite{activitynet,fcvid,kinetics-sounds}, various 2DCNN and 3DCNN based video networks have been proposed in video recognition \cite{twostream,c3d}, showing great performance.
Recently, various adaptive methods have been developed to surmount the limitation of the complexity and computational cost of the models for efficient recognition.
Specifically, current adaptive methods mainly focus on designing lightweight structure \cite{tsm,tinynetwork,mvfnet,xu2022hierarchical,tdn,x3d,teinet}, sampling salient frames/clips \cite{multiagent,adaframe,adamml,tsqnet,nsnet}, selecting frame resolution/key area \cite{fastforward,frameglimpse,arnet,adafocus,adafocusv2,adafocusv3} or adaptive network structure \cite{liteeval,adacs,adafuse}.
These adaptive methods mostly follow a two-stage paradigm of \enquote{preview-then-recognition} \cite{afnet}, i.e., the adaptive selection is performed before the frames are sent to the recognizer.
In this work, we combine the two phases of coarse-grained sampling and fine-grained recognition, and propose a unified spatiotemporal modeling paradigm, which only needs to access the raw frame once during inference.  

\subsection{Long-term video modeling}
Unlike snapshots or clips with only several actions across a few seconds, long-untrimmed video generally consists of multiple semantic units of complex activities over a long range of time.
Aiming to efficiently capture long-term memory in long-untrimmed video, one widely used method is to view a video as a whole, then design lightweight structures to cover more frames and model the temporal causality between frames using RNN \cite{lrcn,arnet,adafocus,liteeval} or 3D-CNN \cite{x3d,p3d,r3d,2+1d,longtermc3d,slowfast} on top of frame-level features.
Yet the representation ability of the hidden states in RNNs as well as 3DCNNs leads to the above-mentioned models short-sightedness, limiting the ability to memorize the features over long time spans.
To remedy this, techniques such as long-term memory bank \cite{adaframe,memorybank} and lightweight cross-modal assistant \cite{listentolook,scsampler} are employed to compensate for features forgotten by the model's short-sightedness.
Another straightforward idea is to split the long-untrimmed videos evenly into multiple segments and further infer the video-level category information after processing the different segments separately \cite{tsn,tbn,eco}.
Unlike existing methods, we investigate the properties of semantic units in video, and propose a hierarchical mechanism to efficiently capture and reason about the unit-level and video-level temporal semantics in long-untrimmed videos respectively, which is conducive to scaling to long videos.

\begin{figure*}[htbp]
\centerline{\includegraphics[width=7 in]{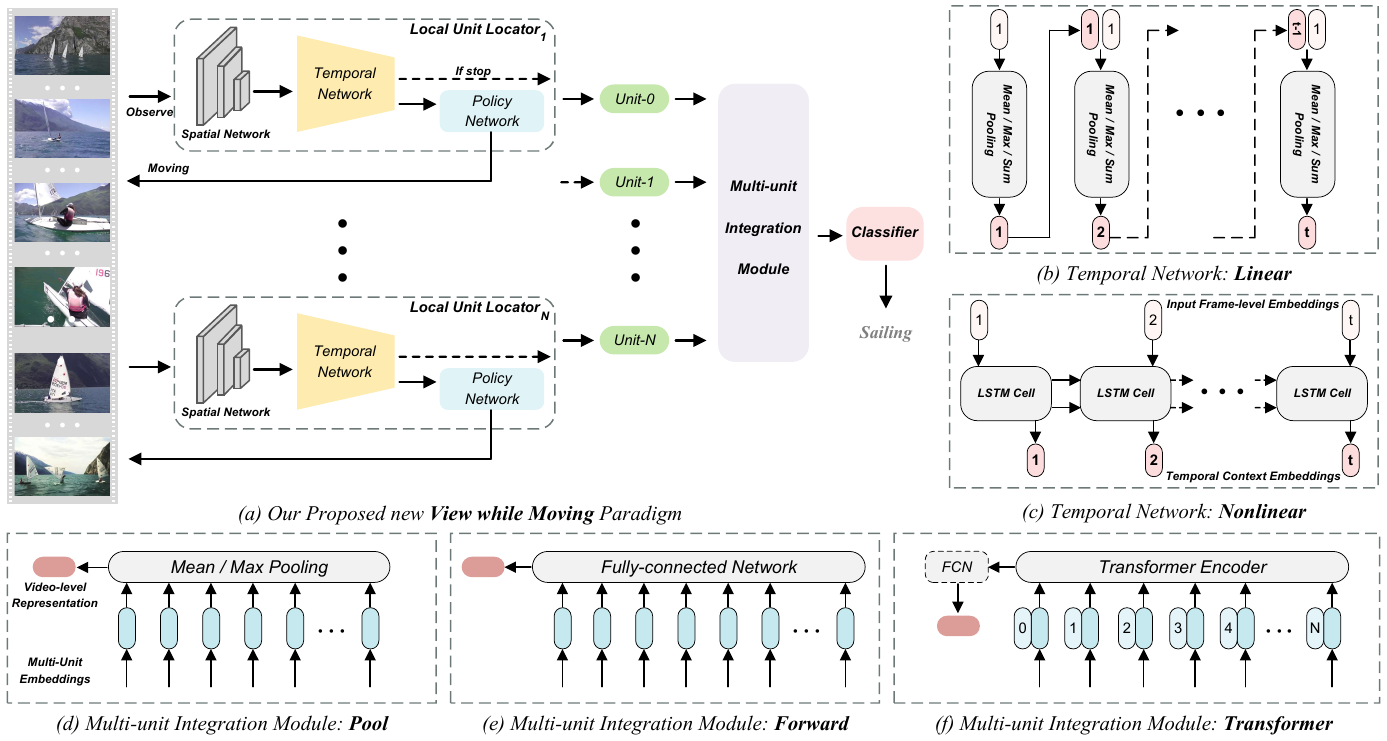}}
\caption{Overall architecture of ViMo.
The whole architecture is illustrated in (a), comprising $N$ \textit{locators} for unit localization and observation, a \textit{multi-unit integration module} for video-level semantics reasoning, and a fully-connected network based \textit{classifier} for categories mapping.
(b) and (c) give \textit{linear} and \textit{nonlinear} options for temporal network of locators.
(d), (e) and (f) present several variants for multi-unit integration module, including \textit{Pool}, \textit{Forward}, and \textit{Transformer}.
Detailed explanations can be found in Section \redpart{\ref{method}}.
Best viewed in color.
}
\label{framework}
\end{figure*}

\section{Method}
\label{method}
In contrast to the existing two-stage paradigm, we seek to save the computation spent on repeated visits to the raw frames, while reasoning about the video-level temporal semantics based on the properties of semantic units in video.
To this end, we propose a novel video recognition paradigm, \enquote{View while Moving}, which adaptively selects temporal salient frames with only one time access to the raw frames, and efficiently observes the local semantic units and reasons about the video-level temporal semantics in long-untrimmed videos with a hierarchical mechanism.

In this section, we first outline the whole framework of the \enquote{View while Moving} paradigm in Section \redpart{\ref{new paradigm}}.
Then we analyze the temporal modeling to further dissect the paradigm in Section \redpart{\ref{temporal modeling analysis}}.
Finally, we present the training strategy and the loss functions used to train up our paradigm in Section \redpart{\ref{training algorithm}}.

\subsection{The New Paradigm "View while Moving"}
\label{new paradigm}

\noindent \textbf{Overview.}
Figure \redpart{\ref{framework}} illustrates an overview of our proposed paradigm "View while Moving".
ViMo has a hierarchical structure, which mainly consists of local unit locators and multi-unit integration module.
For efficient spatiotemporal modeling in long-untrimmed videos, ViMo adaptively localizes the local semantic units of the video, and then integrates the units to reason about video-level temporal semantics for global representation.
Specifically, the local unit locators are uniformly placed along the frames sequence at initial, and individually move to localize and observe local semantic units for local spatiotemporal modeling.
After all locators have stopped, the multi-unit integration module aggregates the unit-level spatiotemporal embeddings to reason about the dynamic long-range contextual dependencies of the whole video.
Finally, a fully-connected network-based classifier estimates the video-level contextual information to output the final prediction.

In general, we combine the two phases of coarse-grained preview and fine-grained recognition, and propose a unified spatiotemporal modeling paradigm, which only needs to access the raw frame once during inference.
The temporal semantics of the video are sliced into two levels, \textit{ i.e.} unit and video, captured and reasoned about separately by a hierarchical framework.
In the following, we describe the components of the paradigm in detail.

\noindent \textbf{Local unit locators} are the outpost that interacts with frames sequence, responsible for localizing and observing the local semantic units in the video. 
To this end, the locator has the ability to model local spatiotemporal features for unit observing and decide where to move adaptively for unit localizing.
Within the locator, there are three components, \textit{i.e.} spatial network $F_S$, temporal network $F_T$, and policy network $F_P$.

Among them, the spatial network $F_S$ parameterized by $\theta_S$ is responsible for encoding the frame-level spatial features, while the temporal network $F_T$ parameterized by $\theta_T$ is in charge of capturing local temporal contextual information.
Frame-level feature extraction frameworks (\textit{e.g.}, ConvNet or Vision Transformer, \textit{etc}.) are all well-suited for the spatial network $F_S$.
To align with the existing work, we adopt a ConvNet-based spatial extractor as $F_S$.
There are several options available for local temporal modeling, such as the linear pool-based method and the nonlinear LSTM, \textit{etc}, as shown in Figure \redpart{\ref{framework}}.
The policy network $F_P$ parameterized by $\theta_P$ is responsible for deciding which frames to observe for adaptive localizing and observing of the semantic unit, consisting of the fully-connected network.
Note that the parameters of the spatial network $F_S$ are shared among all locators.
We defer the details on the architectures of $F_S$, $F_T$, and $F_P$ to Section \redpart{\ref{experiments}}.

Formally, given a video $V$ with a sequence of frames $\{v_1,v_2,...,v_n\}$, $N$ locators are uniformly placed along the frames sequence at initial, expecting to capture multiple local semantic units. At time step $t$, for locator $i$, the common spatial network $F_S$ takes current selected frame $v_{i,t}$ as input and generates the frame-level feature map $e_{i,t}^{s}$:
\begin{equation}
    e_{i,t}^{s}=F_S( v_{i,t} ),
\end{equation}
Then the temporal network $F_{T}^{i}$ processes the current frame-map $e_{{i,t}}^{s}$ with the historical temporal features $h_{i,t-1}$ to produce the current contextual information $h_{i,t}$, which can be denoted as:
\begin{equation}
    h_{i,t}=F_{T}^{i}( e_{{i,t}}^{s}, h_{i,t-1} ),
\end{equation}
On the basis of contextual information $h_{i,t}$, if $t \ge m$ (maximum moving time), the locator stops observing and omits the context $h_{i,t}$ as the local embedding $h_{i}$ of the unit it covers.
If $t\!<m $, the policy network $F_{P}^{i}$ of locator $i$ estimates the current contextual information $h_{i,t}$ and the number of selected frames $n_{i,t}$ to generate a policy distribution $\pi( a_{i,t})$ for sampling decision on whether to early stop or which frame should be observed next, which can be denoted as:
\begin{equation}
    \pi( a_{i,t})=F_{P}^{i}( h_{i,t},n_{i,t} ),
\end{equation}
where $a_{i,t} \in \{ 0,\delta,2\delta,3\delta \} $, and $\delta$ represents the minimum moving stride.

\begin{table*}[t]
    \caption{\textbf{ Ablation studies on hierarchical spatiotemporal modeling mechanism.} Note that the Frame Rate is measured on the basis of 120. The best results are bold-faced, and the default settings of our framework are marked in \colorbox{defaultcolor}{grey}.}
    \label{ablation experiments}
    \begin{subtable}[t]{0.33\textwidth}
        \centering
		\setlength{\tabcolsep}{2pt}
		\renewcommand{\arraystretch}{1.05}
        \caption{Number of local unit locators.}
        \label{Number of patches}
        \begin{tabular}{cccc}
        \toprule
        Locator & Fra. Rat. (\%) & mAP (\%)  & GFLOPs\\
        \midrule
        1&2.5&64.5 & \textbf{15.1}\\
        \df{3}&\df{\textbf{7.1}}& \df{\textbf{82.4}} & \df{38.7}\\
        5&13.5&80.5 & 73.6\\
        8&21.1&79.9 & 116.2\\
        \bottomrule
        \end{tabular}
    \end{subtable}
    \begin{subtable}[t]{0.33\textwidth}
        \centering
		\setlength{\tabcolsep}{2pt}
		\renewcommand{\arraystretch}{1.05}
        \caption{Local spatiotemporal modeling.}
        \label{Local spatiotemporal modeling}
        \begin{tabular}{cccc}
        \toprule
        Variant & Fra. Rat. (\%)  & mAP (\%) & GFLOPs\\
        \midrule
        MeanP & 17.1 & 77.1 & 94.1\\
        MaxP & 18.9 & 77.7 & 104.2\\
        SumP & 16.2 & 76.9 & 84.1\\
        \df{LSTM} & \df{\textbf{7.1}} & \df{\textbf{82.4}} & \df{\textbf{38.7}}\\
        \bottomrule
        \end{tabular}
    \end{subtable}
    \begin{subtable}[t]{0.33\textwidth}
        \centering
		\setlength{\tabcolsep}{2pt}
		\renewcommand{\arraystretch}{1.05}
        \caption{Global semantics reasoning.}
        \label{Global dependencies modeling}
        \begin{tabular}{cccc}
        \toprule
        Variant & Fra. Rat. (\%)  & mAP (\%) & GFLOPs\\
        \midrule
        MeanP & 18.4 & 70.9 & 100.6\\
        MaxP & 33.1 & 71.1 & 181.4\\
        Forward & 19.3 & 72.0 & 105.9\\
        \df{Transformer} & \df{\textbf{7.1}} & \df{\textbf{82.4}} & \df{\textbf{38.7}}\\
        \bottomrule
        \end{tabular}
    \end{subtable}
\end{table*}

\noindent \textbf{Multi-unit integration module} is a video-level global semantics reasoner parameterized by $\theta_M$, which aggregates the multiple unit-level spatiotemporal embeddings into a consistent representation.

Specifically, given embeddings set of units $\{h_1,h_2,...,h_N\}$, the multi-unit integration module $F_M$ captures the dynamic long-range contextual dependencies among all elements in the set to capture global semantics $G_v$, which can be denoted as:
\begin{equation}
    G_v=F_M ( h_1,h_2,...,h_N ),
\end{equation}

A variety of sequence aggregation methods are applicable for the multi-unit integration module $F_M$, such as avg/max pool, fully-connected network, self-attention, \textit{etc}, as shown in Figure \redpart{\ref{framework}}.

\subsection{Temporal Modeling Analysis}
\label{temporal modeling analysis}
Unlike snapshots or clips with only a few actions, the long-untrimmed video consists of multiple semantic units.
Various units describe multiple local events along time that together constitute the complex activity in the video.
Therefore, inspired by human cognition, we try to incorporate this property of semantic units into the workflow of the video recognition model.

\textbf{Hierarchical mechanism.} To this end, we propose a hierarchical mechanism, which is conducive to scale to long videos.
At the bottom of the framework, multiple local unit locators interact with the frames sequence, locating and summarizing the covered semantic units.
At the upper level, the multi-unit integration module $F_M$ aggregates the multiple embeddings of the semantic units to reason about video-level semantics.
By adjusting the number of locators, our hierarchical mechanism enables easy adaptation to adapt videos in various lengths.
As shown in Section \redpart{\ref{experiments}}, this mechanism achieves competitive results not only in long-untrimmed videos, but also performs well in short-trimmed videos.

\textbf{CTDE framework.} Since the content and distribution of semantic units are video-distinct, the locators are able to adaptively make decisions about where to move and when to stop observing, on the basis of policy network $F_P$.
Considering that the semantic units are not independent of each other, there are temporal dependencies and redundancies among them, which affects the localizing and observing of the units.
Therefore, we adopt a centralized training and decentralized execution (CTDE) framework for policy networks of all locators, similar to that of MADDPG \cite{maddpg}.
Locators only have a local field of view and interact independently with frames during inference.
During training, policy networks are trained jointly with centralized critics by policy gradient, to learn how to model the overall semantics and others’ sampling strategies.
More details of the CTDE framework are deferred to the Appendix.

\subsection{Training Algorithm}
\label{training algorithm}
During training, due to the non-differentiability of localizing local semantic units, policy networks of the locators learn with policy gradient, being incompatible with the standard back-propagation.
Therefore, to ensure the components learn properly, a three-stage training algorithm \cite{adafocus} is introduced, \textit{i.e.} initialization and warming-up, policy-learning, and fine-tuning.

\noindent \textbf{Stage I: initialization and warming-up.}
In the first stage, the spatial network $F_S$, temporal network $F_T$, and multi-unit integration module $F_M$ learn to extract the spatial and temporal features through a stochastic sampling policy for warming-up.
For initialization, $F_S$ adopts the pre-trained weights, while $F_T$ and $F_M$ are all trained from scratch.

Specifically, we temporarily freeze the policy networks of all locators, while leveraging a random sampling policy to warm up the rest components.
Each evenly placed locator randomly views 25\% of the frames between it and the next, to model the local semantics of the unit it covers.
The standard back-propagation is used to optimize the backbone network with cross-entropy loss as:
\begin{equation}
    \begin{split}
    \textnormal{Min} & \underset{\theta _{b}}{\textnormal{im}} \textnormal{ize}  E_{( \tilde{v},y ) \sim V}[ -y\log ( \mathcal{P}_{\tilde{v}} ) ],\\
    & \tilde{v} \sim \textnormal{RandomSample}\left( v \right),
    \end{split}
\end{equation}
where $\theta _{b} = \{\theta _{S},\theta _{T},\theta _{M}\}$, $p_v$ and $y$ represents the predicted and ground-truth label of video $v$ in dataset $V$, respectively. 

\noindent \textbf{Stage II: policy-learning.}
In this stage, the locators learn to adaptively select unit semantic-relevant salient frames for localizing the local semantic units.
We freeze the spatiotemporal modeling networks trained in stage I and leverage policy gradient to optimize the policy networks.
To this end, we design a multi-objective reward function of efficiency and accuracy, to guide the policy networks on how to efficiently localize salient semantic frames of units.
For more robust policy learning, we adopt the discrete soft actor-critic \cite{discretesac} for all locators, which use target network as well as double network tricks in the critic network to alleviate the overestimation of q-values for actions during training.

Specifically, at time step $t$, given the total number of frames selected by all locators $N_t$, the reward can be generated as follows for all generators:
\begin{equation}
r_{t}=(P_{t}^{gt}-P_{t-1}^{gt})-\lambda N_t,
\end{equation}
where $P_{t}^{gt}$ denotes the probability corresponding to the ground-truth label in the prediction distribution at time step $t$, and $\lambda$ is a trade-off factor of precision and efficiency.

More details of the structure and optimization of policy networks are deferred to the Appendix.

\noindent \textbf{Stage III: fine-tuning.}
At the last stage, we alternately continue the spatiotemporal learning in stage I and the policy learning in stage II to fine-tune the whole framework.
This facilitates better integration of the components trained in the first two stages to enhance the overall performance.

\section{Experiments}
\label{experiments}
In this section, we conduct comprehensive exploratory studies and comparison experiments for additional insights on three video benchmarks, showing that ViMo yields improved efficiency and accuracy trade-offs in both long-untrimmed and short-trimmed video recognition.

\subsection{Experimental Setup}
\textbf{Datasets.}
We evaluate the performance of our paradigm on three video benchmarks, namely ActivityNet \cite{activitynet}, FCVID \cite{fcvid} and Kinetics-Sounds \cite{kinetics-sounds}. Details of datasets are deferred to the Appendix.

\noindent \textbf{Evaluation metrics.}
To measure the recognition performance, we report the video-level mean average precision (mAP) or top-1 accuracy. 
We also record the giga floating-point operations (GFLOPs) to measure the inference cost of the model.

\noindent \textbf{Implementation details.} For the spatial network $F_S$, we use an ImageNet \cite{imagenet} pre-trained ResNet-50 \cite{resnet} to extract the frame-level feature and output in 1024 dimensions for all locators. 
For temporal network $F_T$: \textit{LSTM}, the hidden dimension of the LSTM \cite{lstm} is set to 256. 
Policy network $F_P$ and critic network $F_C$ of all locators consist of 4 and 5 layers of 512 dimensions respectively. 
For the multi-unit integration module $F_M$: \textit{Transformer}, we stack 8 transformer encoders \cite{transformer}, each with 4 attention heads and 256 dimensions of input embedding.
For the multi-unit integration module $F_M$: forward, we leverage a two-layer fully-connected network of 512 dimensions to aggregate all unit-level embeddings.
We implement the framework with PyTorch 1.11.0.

Due to the limited space, pre-processing and training details are deferred to the Appendix.

\begin{table}[t]
    \caption{Ablations on semantic unit observing strategy.}
    \label{Local Sampling strategy}
    \centering
    \setlength{\tabcolsep}{2pt}
    \renewcommand{\arraystretch}{1.05}
    \begin{tabular}{cccc}
    \toprule
    Strategy & Frame Rate (\%)& mAP (\%) & GFLOPs\\
    \midrule
    All & 100 & 81.7 & 544.1 \\
    \multirow{2}{*}{Uniform} & 25 & 81.1 & 136.6 \\
    & 50 & 81.4 & 272.3 \\
    \multirow{2}{*}{Random} & 25 & 80.3 & 136.6 \\
    & 50 & 81.6 & 272.3 \\
    \df{Adaptive} & \df{\textbf{7.1}} &\df{\textbf{82.4}} & \df{\textbf{38.7}} \\
    \bottomrule
    \end{tabular}
\end{table}

\begin{table}[t]
    \caption{Trade-off factor $\lambda$ of accuracy and efficiency.}
    \label{lambda}
    \centering
    \setlength{\tabcolsep}{2pt}
    \renewcommand{\arraystretch}{1.05}
    \begin{tabular}{ccc}
    \toprule
    Trade-off factor $\lambda$ & mAP (\%) & GFLOPs\\
    \midrule
    0.05 & 81.1 & 39.1 \\
    \df{0.1} & \df{\textbf{82.4}} & \df{38.7} \\
    0.15 & 80.7 & 37.7 \\
    0.2 & 80.9 & \textbf{37.4}\\
    \bottomrule
    \end{tabular}
\end{table}

\subsection{Exploratory Studies}
In this section, we inspect various aspects of our proposed ViMo with comprehensive ablation studies on components and parameters. 
Several intriguing properties are observed. 
All ablations are conducted on ActivityNet.

\noindent\textbf{Different number of locators. }
Local unit locators, at the bottom of ViMo, are the outpost that interacts with frames sequence to localize and observe the local semantic units.
In our setup, each locator is responsible for observing the semantics of the unit around it, outputting a unit-level embedding.
It means that the locators are one-to-one matching the semantic units to be observed.
Therefore, the number of locators determines the size of the embedding set of units.
In theoretical terms, it is preferably equal to the number of semantic units in the video.
Interestingly, as shown in Table \redpart{\ref{Number of patches}}, the experimental results are in line with our expected assumptions that three semantic units are sufficient for videos of several minutes duration, yielding excellent trade-offs.
Therefore, we adopt three local unit locators as default.

\noindent\textbf{Local and global spatiotemporal modeling ablations. }
Table \redpart{\ref{Local spatiotemporal modeling}}, \redpart{\ref{Global dependencies modeling}} and \redpart{\ref{Local Sampling strategy}} studies the design of hierarchical spatiotemporal modeling. 

\begin{table}[t]
    \caption{Sampling action space $\varOmega$ of locators.}
    \label{action space}
    \centering
    \setlength{\tabcolsep}{2pt}
    \renewcommand{\arraystretch}{1.05}
    \begin{tabular}{ccc}
        \toprule
        Action space $\varOmega$ & mAP (\%) & GFLOPs\\
        \midrule
        \{0, 1, 2, 3\} & 78.7 & 45.3 \\
        \{0, 2, 4, 6\} & 80.5 & 44.9 \\
        \df{\{0, 3, 6, 9\}} & \df{\textbf{82.4}} & \df{38.7} \\
        \{0, 4, 8, 12\} & 81.9 & 40.3  \\
        \{0, 5, 10, 15\} & 80.8 & \textbf{34.7}  \\
        \bottomrule
    \end{tabular}
\end{table}

\begin{table}[t]
    \caption{\textbf{Ablation on training algorithm.}}
    \label{training stages}
    \centering 
    \begin{adjustbox}{max width=0.48\textwidth}
        \begin{tabular}{cccc}
            \toprule
            Training stage & ini.\&war. & pol.-lea. & fin.-tun.\\
            \midrule
            mAP (\%) & 80.3 & 81.1 & \textbf{82.4} \\
            GFLOPs  & 136.6 & 46.2 & \textbf{38.7}\\
            \bottomrule
        \end{tabular}
    \end{adjustbox}
\end{table}

For local spatiotemporal modeling, We designed linear and non-linear aggregation methods.
\textit{MeanP}, \textit{MaxP} and \textit{SumP} are commonly used linear aggregation methods that output the average, maximum, and sum of embeddings of multiple frames respectively.
\textit{LSTM} is a classical nonlinear sequence modeling method with gates to control hidden and cell states for previous memory storage and interaction.
As shown in Table \redpart{\ref{Local spatiotemporal modeling}}, we find that the nonlinear method LSTM has a significant advantage in performance for local spatiotemporal modeling.
The performance of the linear method is basically the same, but \textit{MaxP} requires significantly more consumption than the other two methods.
We believe this is due to the fact that \textit{MaxP} discards excessive information, causing the model to observe more frames.
we adopt \textit{LSTM} as default since it has the best results.

\begin{table*}[t]
    \caption{\textbf{Comparison with the state-of-the-art methods on ActivityNet and FCVID.} MN, RN, EN, and Trf denote MobileNet, ResNet, EfficientNet, and Transformer respectively.  
    ViMo yields improved trade-offs in long-untrimmed video recognition.}
    \label{Comparison on activitynet and fcvid}
    \centering
    \begin{adjustbox}{max width=\textwidth}
        \begin{tabular}{ccccccc}
        \toprule
        
        \multirow{2}{*}{Method} & \multicolumn{2}{c}{Backbones}&\multicolumn{2}{c}{ActivityNet} & \multicolumn{2}{c}{FCVID}  \\
        \cmidrule{2-7}
        & Spatial & Temporal & mAP (\%) & GFLOPs  & mAP(\%) & GFLOPs \\
        \midrule
        SCSampler \cite{scsampler} & MN2+RN50 & AvgPool
        &  72.9  & 42.0  & 81.0 & 42.0  \\ 
        AdaFrame \cite{adaframe} & MN2+RN101 & LSTM
        &  71.5  & 78.7 & 80.2 & 75.1 \\
        LiteEval \cite{liteeval} & MN2+RN101 & Dual-LSTM
        &  72.7  & 95.1  & 80.0 & 94.3  \\ 
        ListenToLook (IA|IA) \cite{listentolook} & MN2+RN101 & LSTM
        &  72.3 & 81.4  & - & -  \\ 
        AR-Net \cite{arnet} & MN2+RN50 & LSTM
        &  73.8  & 33.5  & 81.3 & 35.1  \\ 
        AdaMML \cite{adamml} & MN2+RN50 & LSTM
        &  73.9  & 94.0  & 85.8 & 93.9  \\ 
        VideoIQ \cite{videoiq} & MN2+RN50 & LSTM
        &  74.8  & 28.1  & 82.7 & 27.0  \\
        AdaFocus \cite{adafocus} & MN2+RN50 & LSTM
        &  75.0  & 26.6  & 83.4 & 26.6  \\   
        TSQNet \cite{tsqnet} & MN2+ENB0+RN50 & Transformer
        &  76.6  & 26.1  & 83.5 & 26.2  \\ 
        NSNet \cite{nsnet} & MN2+RN50 & Transformer
        &  76.8  & 26.0 & 83.9 & \textbf{26.0}  \\ 
        AdaFocus v2 \cite{adafocusv2} & MN2+RN50 & LSTM
        &  79.0  & 27.0  & 85.0 & 27.0  \\ 
        \cmidrule{1-7}
        ViMo($m=3$) & ResNet-50& LSTM+Trf
        & 79.4 & \textbf{24.1} & 84.8 & 26.7 \\
        ViMo($m=4$) & ResNet-50& LSTM+Trf
        & \textbf{82.4} & 38.7 & \textbf{86.4} & 36.4 \\
        \bottomrule
        \end{tabular}
    \end{adjustbox}
\end{table*}

\begin{table}
    \begin{minipage}[c]{0.48\textwidth}
        \captionof{table}{\textbf{Comparison of practical efficiency on ActivityNet.}}
        \label{practicalefficiency}
        \centering
		\setlength{\tabcolsep}{2pt}{
		\renewcommand{\arraystretch}{0.9}
        \begin{tabular}{c|cccc}
            \toprule
            \multirow{2}*{Method}& \multirow{2}*{mAP (\%)} & \multirow{2}*{GFLOPs} & Latency & Throughput\\
             & & & (bs=1,ms) & (bs=32, vid/sec)\\
            \midrule
            AdaFocus \cite{adafocus}  & 75.0 & 26.6 & 181.8 & 73.8\\
            FrameExit \cite{frameexit}  & 76.1 & \textbf{26.1} & 102.0 & -\\
            \midrule
            ViMo & \textbf{82.4} & 38.7 & \textbf{46.1} & \textbf{129.5} \\
            \bottomrule
        \end{tabular}}
    \end{minipage}
\end{table}

As shown in Table \redpart{\ref{Global dependencies modeling}}, we measure several commonly used video-level aggregation methods, \textit{i.e.} \textit{MeanP}, \textit{MaxP}, \textit{Forward} and \textit{Transformer}, \textit{etc}.
\textit{Forward} utilizes a two-layer, fully-connected network to map the interaction and aggregation of all semantic units.
\textit{Transformer} attends all semantic units based on MHSA, which allows information passing among features of all units, enabling video-level semantics integration and reasoning.
The performance of several aggregation methods is similar to that of local spatiotemporal modeling, but the gaps are even wider.
\textit{Transformer} demonstrates powerful global semantics modeling capabilities, accompanied by remarkably less computation.
We believe this is because methods with enhanced reasoning capability could rely on limited frames for global semantics modeling, saving more computation while achieving better accuracy.
We utilize \textit{Transformer} as default, which performs well trade-offs of accuracy and efficiency.

\noindent\textbf{Ablations on semantic unit localizing strategy. }
In our hierarchical mechanism, locators are employed to localize and observe local semantic units.
Since the content and distribution of semantic units are video-distinct, we design a fully-connected network-based policy network within the locator to adaptively make decisions about where to move and when to stop observing, called \textit{Adaptive}.
To verify its effectiveness, we ablate the local sampling mechanism, via adopting multiple sampling strategies and sampling ratios, including \textit{Uniform} and \textit{Random} sampling strategies.
The results are presented in Table \redpart{\ref{Local Sampling strategy}}.
\textit{Adaptive} localizing strategy significantly reduces the inference consumption while enhancing the accuracy.
Deterministic strategies compromise recognition accuracy while introducing additional computation consumption.
Interestingly, both in \textit{Uniform} and \textit{Random} sampling, observing more frames leads to better recognition accuracy, but the improvement is not cost-effective with double computation cost.
Overall, the \textit{Adaptive} localizing strategy contributes to the efficient localization of semantic units and the elimination of redundant noise.

\begin{figure}[ht]
    \centerline{\includegraphics[width=0.45\textwidth]{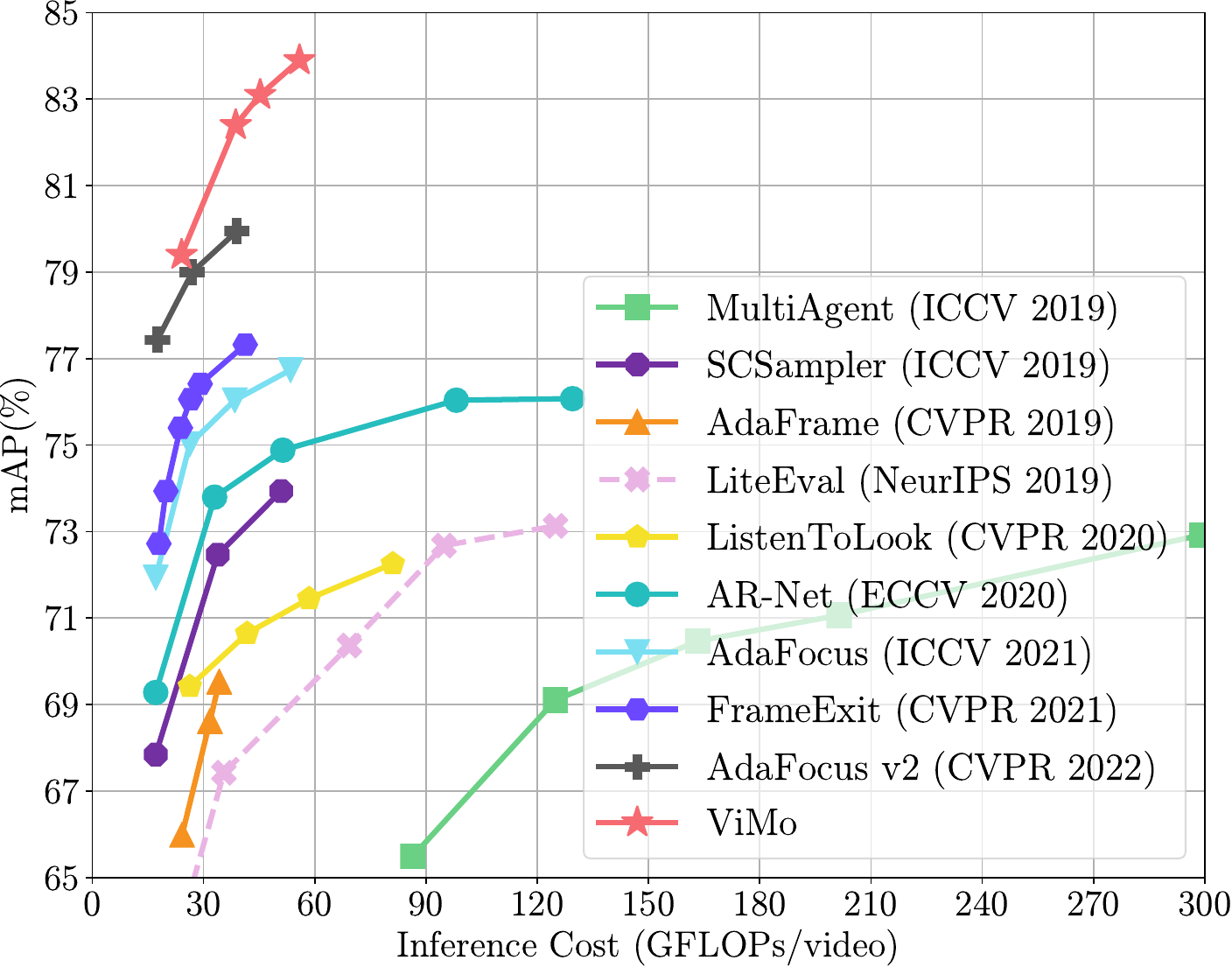}}
    \caption{\textbf{Comparisons with the state-of-the-art methods on ActivityNet.}
    Our method is implemented with the maximum moving times $m\!\in\!$ $\{ 3,4,5,6 \}$.
    Our paradigm achieves competitive performance in terms of mAP as well as GFLOPs.}
    \label{comparison on activitynet(pic)}
\end{figure}

\begin{figure*}[t]
    \centering
    \includegraphics[width=\textwidth]{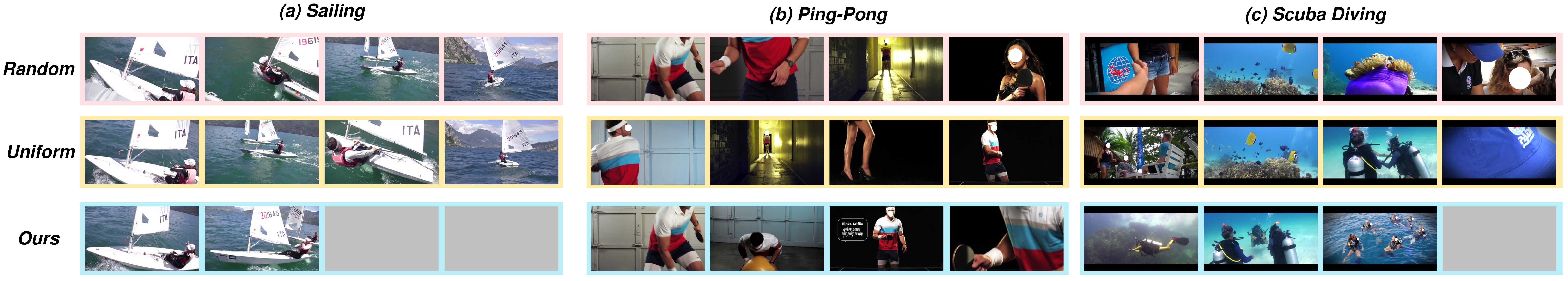}
    \caption{\textbf{Visualization of the random, uniform and adaptive local sampling results by the second locator of our ViMo.} The \colorbox{defaultcolor}{grey} frames indicate that the generator stops sampling and no video frames are observed.
    Please zoom in for more details.
    }
    \label{qualitiveresult}
\end{figure*}

\noindent\textbf{Ablations on policy learning. }
During policy-learning, $\lambda$ and $\delta$ are vital parameters, controlling the trade-off ratio of the locators and the scope of local modeling, respectively.
As shown in Table \redpart{\ref{lambda}}, either greedy or cautious locators could damage accuracy or efficiency.
Greedy locators observe more frames, but do not lead to performance improvement.
Cautious locators reduce computation consumption, but performance decreases as well.
An appropriate ratio $\lambda$ is essential for the trade-offs of accuracy and efficiency.
The same is true for the sampling action space $\varOmega$ in Table \redpart{\ref{action space}}.
Action space with a long step size reduces the amount of computation, but also compromises recognition accuracy, while action space with a short step size damages both efficiency and accuracy.
We believe that the excessive action space results in the poor locality of locators, compromising the hierarchical spatiotemporal modeling, while insufficient action space limits the locators' field of view.
Therefore, we use 0.1 and 3 as defaults for $\lambda$ and $\delta$.

\noindent\textbf{Ablations on training algorithm. }
To ensure the components learn properly, we propose a 3-stage algorithm for training, detailed in Section \redpart{\ref{training algorithm}}. 
In this section, we ablate the training stages, \textit{i.e.} initialization and warming-up, policy-learning, and fine-tuning.
As shown in Table \redpart{\ref{training stages}}, we report the mAP and GFLOPs after each training stage.
During initialization and warming-up, the framework learns the basic representation capabilities of how to model the temporal semantics.
After policy-learning, the adaptive sampling significantly reduces the computation cost of the model, while iterative fine-tuning further optimizes the overall performance of the framework.
Overall, any stage of the training algorithm is essential for both accuracy and efficiency improvements.

\subsection{Comparison with the State-of-the-arts}
\textbf{Long-untrimmed video recognition. }
As shown in Table \redpart{\ref{Comparison on activitynet and fcvid}} and Figure \redpart{\ref{comparison on activitynet(pic)}},
we compare ViMo with state-of-the-art methods on ActivityNet and FCVID, with maximum moving time $m\!\in\!$ $\{ 3,4,5,6 \}$.
It can be observed that ViMo achieves well trade-offs of accuracy and efficiency.
Compared with recent state-of-the-arts, ViMo (m=4) outperforms all the compared methods, with the best performance of 82.4\% and 86.4\% in mAP on ActivityNet and FCVID respectively.
With the same cost of around 27.0 in GFLOPs, ViMo (m=3) achieves competitive results of 79.4\% and 84.8\% in mAP.
We attribute these improvements to the unified \enquote{View while Moving} paradigm and its hierarchical spatiotemporal modeling structure.
As shown in Figure \redpart{\ref{comparison on activitynet(pic)}}, we also can observe that our ViMo can be enhanced significantly with less additional computation cost, compared with existing state-of-the-arts.
These observations indicate that our proposed paradigm, ViMo, is a promising alternative to efficient long-untrimmed video recognition.

\begin{table}[t]
    \caption{\textbf{Comparisons with the state-of-the-arts on Kinetics-Sounds.} ViMo outperforms LiteEval and AdaMML by a large margin in both accuracy and GFLOPs.}
    \label{state-of-the-art on kinetics-sounds}
    \centering
    \begin{adjustbox}{max width=0.48\textwidth}
        \begin{tabular}{ccccc}
            \toprule
            \multirow{2}{*}{Method} & \multicolumn{2}{c}{Backbones}&\multirow{2}{*}{Accuracy (\%)} & \multirow{2}{*}{GFLOPs}  \\
            \cmidrule{2-3}
            & Spatial & Temporal & & \\
            \midrule
            LiteEval \cite{liteeval} & MN2+RN101 & Dual LSTM
            &  72.0 & 104.1 \\
            AdaMML \cite{adamml} & MN2+RN50 & LSTM
            &  89.1 & 142.0 \\
            \cmidrule{1-5}
            ViMo($m=3$) & ResNet-50 & LSTM+Trf
            & 91.8 & \textbf{23.7} \\
            ViMo($m=4$) & ResNet-50 & LSTM+Trf
            & \textbf{92.4} & 33.8 \\
            \bottomrule
        \end{tabular}
    \end{adjustbox}
\end{table}

\noindent\textbf{Comparison of practical efficiency. }
We further evaluate the practical efficiency of inference on ActivityNet, with latency (batchsize = 1) and throughput (batchsize = 32). 
All the tests are performed on the same hardware of one NVIDIA RTX 3090 GPU.
Compared with recent state-of-the-art methods AdaFocus \cite{adafocus} and FrameExit \cite{frameexit} in Table \redpart{\ref{practicalefficiency}}, our proposed ViMo achieves good latency as well as throughput while yielding significant performance improvement.

\noindent\textbf{Additional comparison results. }
To simultaneously validate the performance of our model on short-trimmed videos, experiments are further conducted on Kinetics-Sounds with snapshots of several seconds.
As shown in Table \redpart{\ref{state-of-the-art on kinetics-sounds}}, we compare ViMo with LiteEval \cite{liteeval} and AdaMML \cite{adamml} on the top-1 accuracy and GFLOPs. 
Overall, our proposed new unified paradigm ViMo not only yields improved efficiency and accuracy trade-offs in long-untrimmed videos, but also performs well in short-trimmed snapshots.

\section{Qualitative Results}
Figure \redpart{\ref{qualitiveresult}} shows the selected frames of the second locator, using our proposed ViMo on different sampling strategies across multiple classes (\textit{Sailing}, \textit{Ping-Pong} and \textit{Scuba Diving}).
For \textit{Sailing}, adaptive sampling selects only two salient frames of local semantic unit \enquote{Controlling the sailboat} for observation, significantly eliminating redundancy.
In contrast, the other two strategies select more frames both related to the subject of the semantic unit but too redundant.
Salient frames of \enquote{Warming-up} in \textit{Ping-Pong} and \enquote{Diving and surfacing} in \textit{Scuba Diving} are selected by adaptive sampling for observation.
The other two strategies introduce a large amount of noise, that almost drowns out the subject of the semantic unit.
In ours, irrelevant noises in \textit{Ping-Pong} and \textit{Scuba Diving} are discarded, while the other two strategies introduce excessive noises.
Therefore, it is observed that, with adaptive sampling, the salient frames related to the semantics of the unit are observed for local spatiotemporal modeling, while irrelevant noises or redundancies are discarded, demonstrating the effectiveness of our paradigm.

\section{Conclusion}
In this paper, we propose a new \enquote{View while Moving} paradigm for efficient long-untrimmed video recognition, which only accesses the raw frame at once during inference.
Extensive experiments on three video benchmarks show that our proposed ViMo not only yields improved efficiency and accuracy trade-offs in long-untrimmed video recognition, but also performs well in short-trimmed videos.
In the future, we plan to continue exploring the interaction and communication between locators in the parallel process and end-to-end training of the whole framework.

\section*{ACKNOWLEDGMENTS}
This work was supported in part by the National Natural Science Foundation of China under Grant 62072048, and in part by Industry-University-Research Innovation Fund of Universities in China under Grant 2021ITA07005.

\balance
\newpage
\bibliographystyle{ACM-Reference-Format}
\bibliography{references}

\appendix

\newpage

\section{Structure of Locator}
\begin{figure}[h]
\centerline{\includegraphics[width=0.48\textwidth]{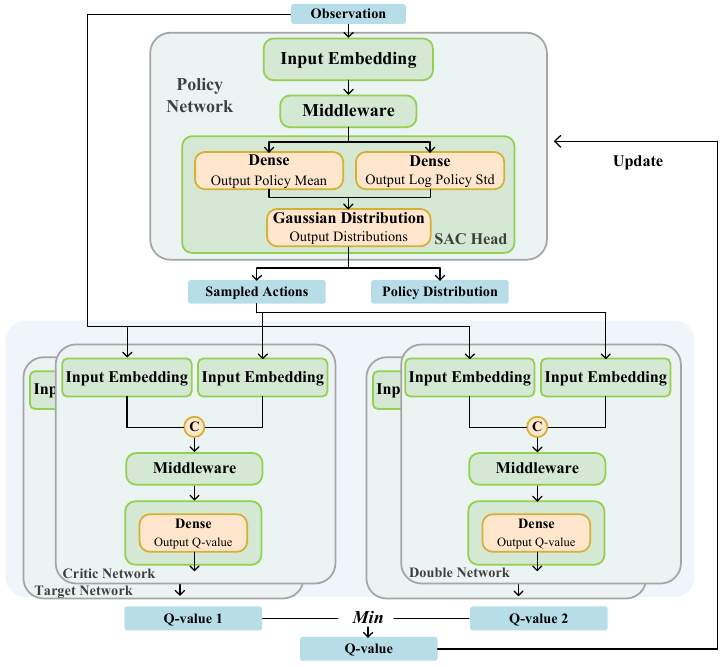}}
\caption{\textbf{Structure of Locator.}
}
\label{locator}
\end{figure}

Within ViMo, we initialize multiple local unit locators to localize and observe the local semantic units of the video.
Since the content and distribution of semantic units are video-distinct, the locators are able to adaptively make decisions about where to move and when to stop observing
As shown in Figure \ref{locator}, each locator contains a policy network $f_p$ for decision-making and a critic network $f_r$ for learning. 
Both the policy and the critic networks consist of fully connected networks.
To alleviate the overestimation of q-values during training, we adopt the architecture of discrete Soft Actor-Critic \cite{discretesac} for all locators, using the double network as well as target network tricks within the critic network.

The double network trick initializes two critic networks with the same structure but different parameters, and the smaller Q-value is used as the final criterion during training.
The target network trick initializes a same-structure and same-parameters target network for each critic network, learning with soft momentum updating, as:
\begin{equation}
    \theta_{target} = \tau*\theta_{local} + (1 - \tau)*\theta_{target},
\end{equation}
where the $\theta_{target}$ and  $\theta_{local}$ denote the parameters of the target network and local critic network respectively, and $\tau$ is a soft factor controlling the updating scale with 0.99 by default.
The target network is only used in the evaluation of the policy network during training.

Specifically, for locator $i$, given sampled tuple $( s_{i,t},a_{i,t},r_t,s_{i,t+1} )$ from replay buffer $D$, the policy loss can be formulated as follow:
\begin{multline}
J_{\pi}( \theta _{P}^{i} ) =E_{s_{i,t} \sim D}[ \pi (s_{i,t} ;{\theta _{P}^{i}}) ^T[ \alpha \log ( \pi( s_{i,t} ; {\theta _{P}^{i}}) )\\
- \underset{j=1,2}{\min} Q( s_{i,t} ; {\theta _{C_j}^{i}}) ] ] ,
\end{multline}
where $Q( \cdot ; {\theta _{C_j}^{i}})$ denotes double critic network $r_j$ of locator $i$.

Also, the loss of critic $Q( \cdot ; {\theta _{C}^{i}})$ can be formulated as follow:
\begin{multline}
J_Q( \theta _{C}^{i} ) =E_{( s_{i,t},a_{i,t},r_t,s_{i,t+1} ) \sim D}[ \frac{1}{2}( Q( s_{i,t};\theta _{C}^{i} ).gather( a_{i,t} )\\
-( r_t+\gamma V( s_{i,t+1} ) ) ) ^2 ] ,
\end{multline}
\begin{multline}
V( s_{i,t+1}) =\pi ( s_{i,t+1};{\theta _{P}^{i}}) ^T[ \underset{j=1,2}{\min} Q( s_{i,t+1};{\bar{\theta}_{C_j}^{i}}),\\
 - \alpha \log ( \pi( s_{i,t+1};{\theta _{P}^{i}}) ) ] ,
\end{multline}
where $Q(\cdot ; {\bar{\theta}_{C_j}^{i}})$ is the target critic network, and $\gamma$ denotes the constant reward discount factor.

During training, temperature $\alpha$ controls the exploration and convergence of the policy learning. 
Specifically, $\alpha$ works as an adaptive scale factor, 
which dynamically adjusts the scale of the action entropy to limit its average above $\bar{H}$.
The loss of $\alpha$ can be formulated as follow:
\begin{equation}
J_{\alpha}( \alpha ) =\pi (  s_{i,t} ; {\theta _{P}^{i}}) ^T[ -\alpha ( \log ( \pi (  s_{i,t} ; {\theta _{P}^{i}}) ) +\bar{H} ) ] ,
\end{equation}
where $\bar{H}$ is a constant vector representing the target entropy.

Note that the critic networks work only during the training, guiding the policy network for learning.
During inference, only the policy network plays a role, estimating the observation and outputting the action distribution.

\section{Learning Flowline of Locators}
Considering the semantic dependencies between local units, the locators are not independent of each other, which affects the localizing and observing of the units.
Therefore, we adopt a centralized training and decentralized execution strategy similar to that of MADDPG \cite{maddpg}.

\begin{figure}[h]
    \centerline{\includegraphics[width=0.4\textwidth]{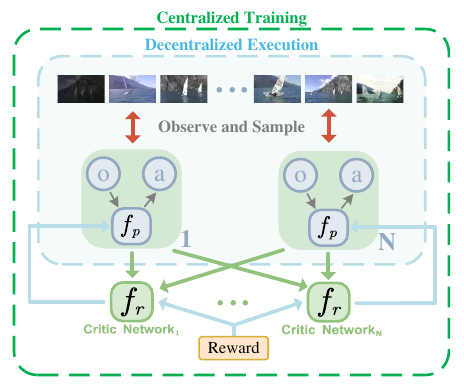}}
    \caption{\textbf{Centralized Training and Decentralized Execution of Locators.}
    }
    \label{ctde}
\end{figure}

As shown in Figure \ref{ctde}, locators are trained jointly by centralized critics, to learn how to model the overall semantics and others' sampling strategies, while independently making decisions during inference.
The critic networks have global observations of observed states and actions of all policy networks for all locators.
In contrast, the policy networks have only local observations and execute independently during inference.

\section{Dataset Details}
We evaluate the performance of our approach on three commonly used video benchmarks, namely ActivityNet-v1.3 \cite{activitynet}, FCVID \cite{fcvid} and Kinetics-Sounds \cite{kinetics-sounds}, as detailed below.

\textbf{ActivityNet-v1.3} mainly contains long-untrimmed videos with an average duration of about 117 seconds in 200 categories.
It has 10,024 videos for training and 4926 videos for validation. 
We use the origin training and validation split in our experiments to train and test our framework.
The dataset is available to download at \redpart{\url{http://activity-net.org/download.html}}.

\textbf{FCVID} mainly contains long-untrimmed videos with an average duration of about 167 seconds across 239 categories.
FCVID covers a wide range of topics such as social activities (e.g., “tailgate party”), procedural actions (e.g., “making cake”), objects (e.g., “panda”), and scenes (e.g., “beach”), etc. 
We use the origin training/testing split in our experiments, containing 45,611 videos and 45,612 videos respectively.
The dataset is available to download at \redpart{\url{https://fvl.fudan.edu.cn/dataset/fcvid/list.htm}}.

\textbf{Kinetics-Sounds} \cite{listentolook} is a subset of Kinetics \cite{kinetics} with 34 action classes, which consists mainly of short-trimmed snapshots with an average duration of about 9.6 seconds. 
According to the original training-validation split, it has 22,521 videos for training and 1,532 videos for testing.
As 3 classes were removed from the Kinetics, we used the remaining 31 classes in our experiments.
The Kinetics dataset is available to download at \redpart{\url{https://deepmind.com/research/open-source/kinetics} and the classes of Kinetics-Sounds is available from}.

\section{Data Pre-processing}
For all datasets, we decode the video into RGB frames at 1fps using FFmpeg with average duration, which is 120s, 170s, and 36s for ActivityNet, FCVID, and Kinetics-Sounds, respectively.
We resize the short side of all frames to 256, then crop them to 224x224 randomly during training while centrally during testing.
In addition, we randomly horizontally flip the video frames for augmentation during training.

\section{Training Details}
During training, we use Adam as the optimizer for all training stages.
At Stage I, the learning rate is 1e-5. The batchsize is 8 for ActivityNet and FCVID, while 10 for Kinetics-Sounds. 
At policy-learning, the learning rates of the policy network, critic network, and alpha are set to 1e-5, 5e-5, and 5e-4, respectively. 
The batchsize is 8 for all datasets. 
At fine-tuning, we alternatively fine-tune the backbone and policy network, switching every 5 epochs.
The learning rates are all set to 1e-5, and the batchsize is 8 for all datasets. 

\section{Initial Frames}
The frames at the initial positions are indeed fused as part of local semantic units, which are not selected adaptively. When all locators are placed uniformly, there is a "cold start" problem before adaptive selection. The locator is not aware of the contextual information of the location to adaptively move, and must observe the frames of the initial location to decide the subsequent movement process. At each observation location, the locator fuses the observed features into contextual information via Temporal Network without preserving the features of the single frames. When stopped, the locator's context information will be emitted directly as the semantics of the unit.

\begin{table}[ht]
    \caption{\textbf{Ablation on initial frames.}}
    \label{initial frame}
    \centering
    \setlength{\tabcolsep}{2pt}
    \renewcommand{\arraystretch}{1.05}
    \begin{tabular}{ccc}
        \toprule
        Initial Frame & mAP & GFLOPs\\
        \midrule
        \df{\textbf{w/}} & \df{\textbf{82.4}} & \df{\textbf{38.7}} \\
        w/o  & 82.1 & 38.7\\
        \bottomrule
    \end{tabular}
\end{table}

In this section, we try to drop the first frame directly after the initial position decision.
As shown in Table \redpart{\ref{initial frame}}, discarding the initial frame directly does not work well.
Direct discarding can not reduce the computational effort of model inference, and results in a loss of information, leading to a decrease in mAP.

\balance

\end{document}